\newif\ifjournal\journalfalse   

\ifjournal
\documentclass{elsart}
\usepackage{latexsym}
\else

\documentclass[12pt,twoside]{article}
\usepackage{latexsym}

\fi

\ifjournal
\def\beq{\begin{equation}}    \def\eeq{\end{equation}}
\def\beqn{\begin{displaymath}}\def\eeqn{\end{displaymath}}
\def\bqa{\begin{eqnarray}}    \def\eqa{\end{eqnarray}}
\def\bqan{\begin{eqnarray*}}  \def\eqan{\end{eqnarray*}}
\else
\def\,{\mskip 3mu} \def\>{\mskip 4mu plus 2mu minus 4mu} \def\;{\mskip 5mu plus 5mu} \def\!{\mskip-3mu}
\def\dispmuskip{\thinmuskip= 3mu plus 0mu minus 2mu \medmuskip=  4mu plus 2mu minus 2mu \thickmuskip=5mu plus 5mu minus 2mu}
\def\textmuskip{\thinmuskip= 0mu                    \medmuskip=  1mu plus 1mu minus 1mu \thickmuskip=2mu plus 3mu minus 1mu}
\textmuskip
\def\eqsp{\vspace{0ex}}
\def\beq{\dispmuskip\eqsp\begin{equation}}    \def\eeq{\eqsp\end{equation}\textmuskip}
\def\beqn{\dispmuskip\eqsp\begin{displaymath}}\def\eeqn{\eqsp\end{displaymath}\textmuskip}
\def\bqa{\dispmuskip\eqsp\begin{eqnarray}}    \def\eqa{\eqsp\end{eqnarray}\textmuskip}
\def\bqan{\dispmuskip\eqsp\begin{eqnarray*}}  \def\eqan{\eqsp\end{eqnarray*}\textmuskip}
\fi

\ifjournal
\def\cal{\mathcal}
\else
\newenvironment{keyword}{\centerline{\bf\small
Keywords}\vspace{-1ex}\begin{quote}\small}{\par\end{quote}\vskip 1ex}
\fi
\newtheorem{theorem}{Theorem}
\newtheorem{corollary}[theorem]{Corollary}
\newtheorem{lemma}[theorem]{Lemma}
\newtheorem{definition}[theorem]{Definition}
\newtheorem{tablex}[theorem]{Table}
\newtheorem{figurex}[equation]{Figure}

\def\ftheorem#1#2#3{\begin{theorem}[#2]\label{#1} #3 \end{theorem} }

\def\flemma#1#2#3{\begin{lemma}[#2]\label{#1} #3 \end{lemma} }
\def\fdefinition#1#2#3{\begin{definition}[#2]\label{#1} #3 \end{definition} }
\def\ftablex#1#2#3{\begin{tablex}[#2]\label{#1} #3 \end{tablex} }

\ifjournal
\def\paradot#1{{\itshape{#1.}}}
\def\paranodot#1{{\itshape{#1}}}
\else
\def\myparskip{\vspace{1.5ex plus 0.5ex minus 0.5ex}\noindent}
\def\paradot#1{\myparskip{\bfseries\boldmath{#1.}}}
\def\paranodot#1{\myparskip{\bfseries\boldmath{#1}}}
\fi
\def\toinfty#1{\stackrel{#1\to\infty}{\longrightarrow}}
\def\nq{\hspace{-1em}}
\def\qed{\hspace*{\fill}$\Box\quad$}
\def\odn{{\textstyle{1\over n}}}
\def\odt{{\textstyle{1\over 2}}}
\def\odf{{\textstyle{1\over 4}}}
\def\eps{\varepsilon}                   
\def\epstr{\epsilon}                    
\def\qmbox#1{{\quad\mbox{#1}\quad}}
\def\geqm{\unrhd}
\def\ngeqm{{\not\unrhd}}

\def\l{{\ell}}                          
\def\M{{\cal M}}                        
\def\X{{\cal X}}                        

\def\E{{\bf E}}                         
\def\P{{\bf P}}                         
\def\B{\{0,1\}}                        
\def\Km{K\!m}
\def\MM{M}                              
\def\th{\theta}
\def\e{{\rm e}}                        
\def\SetN{I\!\!N} \def\SetQ{I\!\!\!Q} \def\SetR{I\!\!R} \def\SetZ{Z\!\!\!Z}
\def\lb{\log}
\def\sumprime{\mathop{{\sum\nolimits'}}}
\def\text#1{\mbox{\scriptsize{#1}}}    

\begin{document}

\ifjournal

\begin{frontmatter}
\title{On Generalized Computable Universal Priors and their Convergence}
\author{Marcus Hutter}
\address{IDSIA, Galleria 2, CH-6928 Manno-Lugano, Switzerland \\
marcus@idsia.ch \hspace{9ex} http://www.idsia.ch/$^{_{_\sim}}\!$marcus}

\thanks{A preliminary version appeared
  in the proceedings of the ALT 2003 conference \cite{Hutter:03unipriors}.
  This work was supported by SNF grant 2000-61847.00 to J\"urgen Schmidhuber.}

\else

\title{\vskip -25mm\normalsize\sc Technical Report \hfill IDSIA-05-05
\vskip 2mm\bf\LARGE\hrule height5pt \vskip 3mm
\sc On Generalized Computable Universal Priors and their Convergence%
\thanks{A preliminary version appeared
  in the proceedings of the ALT 2003 conference \cite{Hutter:03unipriors}.\newline
  \hspace*{4ex}This work was supported by SNF grant 2000-61847.00 to J\"urgen Schmidhuber.}
\vskip 2mm \hrule height2pt \vskip 5mm}
\author{{\bf Marcus Hutter}\\[3mm]
\normalsize IDSIA, Galleria 2, CH-6928\ Manno-Lugano, Switzerland\\
\normalsize marcus@idsia.ch \hspace{8.5ex} http://www.idsia.ch/$^{_{_\sim}}\!$marcus}
\date{11 March 2005}
\maketitle

\fi

\begin{abstract}
\noindent Solomonoff unified Occam's razor and Epicurus' principle
of multiple explanations to one elegant, formal, universal theory
of inductive inference, which initiated the field of algorithmic
information theory. His central result is that the posterior of
the universal semimeasure $\MM$ converges rapidly to the true
sequence generating posterior $\mu$, if the latter is computable.
Hence, $M$ is eligible as a universal predictor in case of unknown
$\mu$. The first part of the paper investigates the existence and
convergence of computable universal (semi)measures for a hierarchy
of computability classes: recursive, estimable, enumerable, and
approximable. For instance, $\MM$ is known to be enumerable, but
not estimable, and to dominate all enumerable semimeasures. We
present proofs for discrete and continuous semimeasures. The
second part investigates more closely the types of convergence,
possibly implied by universality: in difference and in ratio, with
probability 1, in mean sum, and for Martin-L{\"o}f random
sequences. We introduce a generalized concept of randomness for
individual sequences and use it to exhibit difficulties regarding
these issues. In particular, we show that convergence fails
(holds) on generalized-random sequences in gappy (dense) Bernoulli
classes.
\end{abstract}


\ifjournal\end{frontmatter}\else\pagebreak\fi

\section{Introduction}\label{secIntro}

All induction problems can be phrased as sequence prediction
tasks. This is, for instance, obvious for time-series prediction,
but also includes classification tasks. Having observed data $x_t$
at times $t<n$, the task is to predict the $t$-th symbol $x_t$ from
sequence $x=x_1...x_{t-1}$.
The key concept to attack general induction problems is {\em
Occam's razor} (simplicity) principle, which says that ``{\em
Entities should not be multiplied beyond necessity}.'' and to a
less extent Epicurus' principle of multiple explanations. The
former/latter may be interpreted as to keep the simplest/all
theories consistent with the observations $x_1...x_{t-1}$ and to
use these theories to predict $x_t$.
%
Kolmogorov (and others) defined the complexity of a string as the
length of its shortest description on a universal Turing machine.
The Kolmogorov complexity $K$ is an excellent universal complexity
measure, suitable for quantifying Occam's razor. There is (only)
one disadvantage: $K$ is not computable.

More precisely, a function $f$ is said to be {\em recursive} (or
{\em finitely computable}) if there exists a Turing machine that,
given $x$, computes $f(x)$ and then halts. Some functions are not
recursive but still {\em approximable} (or {\em limit-computable})
in the sense that there is a nonhalting Turing machine with an
infinite ($x$-dependent) output sequence $y_1,y_2,y_3,...$ and
$\lim_{t\to\infty}y_t=f(x)$. If additionally the output sequence
is monotone increasing/decreasing, then $f$ is said to be {\em
lower/upper semicomputable} (or {\em enumerable/co-enumerable}).
Finally we call $f$ {\em estimable} if some Turing machine, given
$x$ and a precision $\eps$, finitely computes an
$\eps$-approximation of $x$.
The major algorithmic property of $K$ is that it is co-enumerable,
but not recursive.

More suitable for predictions is Solomonoff's
\cite{Solomonoff:64,Solomonoff:78} {\em universal prior} $\MM(x)$
defined as the probability that the output of a universal monotone Turing
machine $U$ starts with string $x$ when provided with fair
coin flips on the input tape. $\MM(x)$ is enumerable and roughly
$2^{-K(x)}$, hence implementing Occam's and also Epicurus'
principles.

Assume now that strings $x$ are sampled from a probability
distribution $\mu$, i.e.\ the probability of a string starting
with $x$ shall be $\mu(x)$.
The probability of observing $x_t$ at time $t$, given past
observations $x_1...x_{t-1}$ is
$\mu(x_t|x_1...x_{t-1})=\mu(x_1...x_t)/\mu(x_1...x_{t-1})$.
Solomonoff's \cite{Solomonoff:78} central result is that the
universal posterior
$\MM(x_t|x_1...x_{t-1})=\MM(x_1...x_t)/\MM(x_1...x_{t-1})$
converges rapidly to the true (objective) posterior probability
$\mu(x_t|x_1...x_{t-1})$, if $\mu$ is an estimable measure, hence
$\MM$ can be used for predictions in case of unknown $\mu$.
One representation of $\MM$ is as a $2^{-K(\mu)}$-weighted sum of
{\em all} enumerable ``defective'' probability measures, called
semimeasures.
The (from this representation obvious) dominance $\MM(x) \geq
2^{-K(\mu)}\mu(x)$ for all enumerable $\mu$ is the central
ingredient in the convergence proof.

Dominance and convergence immediately generalize to arbitrary
weighted sums of (semi)measures of some arbitrary countable set
$\M$.
So what is so special about the class of all enumerable
semimeasures $\M_{enum}^{semi}$? The larger we choose $\M$ the
less restrictive is the essential assumption that $\M$ should
contain the true distribution $\mu$.
Why not restrict to the still rather general class of estimable or
recursive (semi)measures? For {\em every} countable
class $\M$ and $\xi_\M(x):=\sum_{\nu\in\M} w_\nu \nu(x)$ with
$w_\nu>0$, the important dominance $\xi_\M(x)\geq w_\nu
\nu(x)\,\forall\nu\in\M$ is satisfied. The question is what
properties $\xi_\M$ possesses. The distinguishing property of
$\M_{enum}^{semi}$ is that $\MM=\xi_{\M_{enum}^{semi}}$ is itself
an element of $\M_{enum}^{semi}$.
On the other hand, for prediction, $\xi_\M\in\M$ is not by itself
an important property. What matters is  whether $\xi_\M$ is
computable (in one of the senses we defined above) to avoid
getting into the (un)realm of non-constructive math.

Our first contribution is to classify the existence of generalized
computable (semi)measures.
From \cite{Zvonkin:70} we know that there is an enumerable
semimeasure (namely $\MM$) that dominates all enumerable
semimeasures in $\M_{enum}^{semi}$. We show that there is {\em no}
estimable semimeasure that dominates all recursive measures (also
mentioned in \cite{Zvonkin:70}), and there is {\em no}
approximable semimeasure that dominates all approximable measures.
From this it follows that for a universal (semi)measure that at
least satisfies the weakest form of computability, namely being
approximable, the largest dominated class among the classes
considered in this work is the class of enumerable semimeasures.
This is the distinguishing property of $\M_{enum}^{semi}$ and
$\MM$.
This investigation was motivated by recent
generalizations of Kolmogorov complexity and Solomonoff's prior by
Schmidhuber \cite{Schmidhuber:00toe,Schmidhuber:02gtm}.

The second contribution is to investigate more closely the types of
convergence, possibly implied by universality: in difference and
in ratio, with probability 1, in mean sum, and for Martin-L{\"o}f
random sequences.
We introduce a generalized concept of randomness for individual
sequences and use it to exhibit difficulties regarding these
issues. More concretely, we consider countable classes $\M$ of
Bernoulli environments and show that $\xi_\M$ converges to $\mu$
on all generalized random sequences if and only if the class is
dense.

\paradot{Contents}
In Section~\ref{secCC} we review various computability concepts
and discuss their relation.
In Section~\ref{secUniM} we define the prefix Kolmogorov
complexity $K$, the concept of (semi)measures, Solomonoff's
universal prior $\MM$, and explain its universality.
Section~\ref{secUSP} summarizes Solomonoff's major convergence
result, discusses general mixture distributions and the important
universality property -- multiplicative dominance.
In Section~\ref{secUSM} we define seven classes of (semi)measures
based on four computability concepts. Each class may or may not
contain a (semi)measures that dominates all elements of another
class. We reduce the analysis of these 49 cases to four basic
cases. Domination (essentially by $\MM$) is known to be true for
two cases. The other two cases do not allow for domination.
In Section~\ref{secConv} we investigate more closely the type of
convergence implied by universality. We summarize the result on
posterior convergence in difference $(\xi-\mu\to 0)$ and improve
the previous result \cite{Li:97} on the convergence in ratio
$\xi/\mu\to 1$ by showing rapid convergence without use
of martingales.
In Section~\ref{secMLconv} we investigate whether convergence for
all Martin-L{\"o}f random sequences could hold. We define a
generalized concept of randomness for individual sequences and use
it to show that proofs based on universality cannot decide this
question.
Section~\ref{secConc} concludes the paper.

\paradot{Notation}
We denote strings of length $n$ over finite alphabet $\X$ by
$x=x_1x_2...x_n$ with $x_t\in\X$ and further abbreviate
$x_{1:n}:=x_1x_2...x_{n-1}x_n$ and $x_{<n}:=x_1... x_{n-1}$,
$\epstr$ for the empty string, $\l(x)$ for the length of string $x$,
and $\omega=x_{1:\infty}$ for infinite sequences.
We write $xy$ for the concatenation of string $x$ with $y$.
%
We abbreviate $\lim_{n\to\infty}[f(n)-g(n)]=0$ by
$f(n)\toinfty{n}g(n)$ and say $f$ converges to $g$, without
implying that $\lim_{n\to\infty}g(n)$ itself exists. We write
$f(x)\geqm  g(x)$ for $g(x)=O(f(x))$, i.e.\ if $\exists c>0:
f(x)\geq c g(x)\forall x$.

\section{Computability Concepts}\label{secCC}
We define several computability concepts weaker than can be captured
by halting Turing machines.

\fdefinition{defCompFunc}{Computable functions}{
We consider functions $f:\SetN\to\SetR$:
\begin{itemize}\ifjournal\parskip=0ex\parsep=0ex\itemsep=0.5ex\fi
\item[]
$\nq f$ is {\em recursive} or {\em finitely computable} {\it iff}
there are Turing machines $T_{1/2}$ with output interpreted as natural
numbers and $f(x)={T_1(x)\over T_2(x)}$,
\item[]
$\nq f$ is {\em approximable} or {\em limit-computable} {\it iff}
$\exists$ recursive $\phi(\cdot,\cdot)$ with
$\lim_{t\to\infty}\phi(x,t)=f(x)$.
\item[]
$\nq f$ is {\em enumerable} or {\em lower semicomputable} {\it
iff} additionally $\phi(x,t)\leq\phi(x,t+1)$.
\item[]
$\nq f$ is {\em co-enumerable} or {\em upper semicomputable} {\it
iff} $[-f]$ is lower semicomputable.
\item[]
$\nq f$ is {\em semicomputable} {\it iff} $f$ is lower- {\it or}
upper semicomputable.
\item[]
$\nq f$ is {\em estimable} {\it iff} $f$ is lower- {\it and} upper
semicomputable.
\end{itemize}
}

\noindent If $f$ is estimable we can finitely compute an
$\eps$-approximation of $f$ by upper and lower semicomputing $f$
and terminating when differing by less than $\eps$. This means
that there is a Turing machine which, given $x$ and $\eps$,
finitely computes $\hat y\in\SetQ$ such that $|\hat y-f(x)|<\eps$.
Moreover it gives an interval estimate $f(x)\in[\hat y-\eps,\hat
y+\eps]$. An estimable integer-valued function is recursive (take
any $\eps<\odt$).
Note that if $f$ is only approximable or semicomputable we can
still come arbitrarily close to $f(x)$ but we cannot devise a
terminating algorithm that produces an $\eps$-approximation. In
the case of lower/upper semicomputability we can at least
finitely compute lower/upper bounds to $f(x)$. In case of
approximability, the weakest computability form, even this
capability is lost.

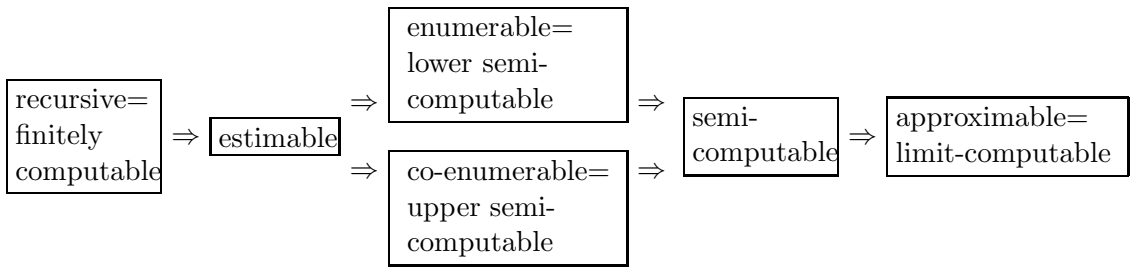
\begin{center}\small
\fbox{\parbox{11ex}{recursive=\\ finitely\\ computable}}
$\Rightarrow$
\fbox{\parbox{9ex}{estimable}}
\parbox{26ex}{\raisebox{-3ex}{$\Rightarrow$} \fbox{
\parbox{17ex}{enumerable=\\lower semi-\\ computable}}
\raisebox{-3ex}{$\Rightarrow$} \\[2ex]
\raisebox{3ex}{$\Rightarrow$} \fbox{
\parbox{17ex}{co-enumerable=\\ upper semi-\\
computable}} \raisebox{3ex}{$\Rightarrow$}}
\fbox{\parbox{11ex}{semi-\\ computable}}
$\Rightarrow$
\fbox{\parbox{18ex}{approximable=\\ limit-computable}}
\end{center}

\noindent What we call {\em estimable/recursive/finitely
computable} is often just called {\em computable}, but it makes
sense to separate the concepts in this work, since finite
computability is conceptually easier and some previous results
have only been proved for this case. Sometimes we us
the word {\em computable} generically for some of the
computability forms of Definition~\ref{defCompFunc}.

\section{The Universal Prior $\MM$}\label{secUniM}

The prefix Kolmogorov complexity $K(x)$ is defined as the length
of the shortest binary (prefix) program $p\in\B^*$ for which a
universal prefix Turing machine $U$ (with binary program tape and
$\X$ary output tape) outputs string $x\in\X^*$, and similarly
$K(x|y)$ in case of side information $y$
\cite{Kolmogorov:65,Levin:74,Gacs:74,Chaitin:75}:
\beqn
  K(x)=\min\{\l(p):U(p)=x\},\qquad
  K(x|y)=\min\{\l(p):U(p,y)=x\}
\eeqn
Solomonoff \cite[Eq.(7)]{Solomonoff:64} defined (earlier) the
closely related quantity, the universal posterior
$\MM(y|x)=M(xy)/M(x)$.
The universal prior $M(x)$ can be defined as the probability that
the output of a universal monotone Turing machine $U$ starts with
$x$ when provided with fair coin flips on the input tape.
Formally, $\MM$ can be defined as
\beq\label{Mdef}
  \MM(x)\;:=\;\sum_{p\;:\;U(p)=x*}\nq 2^{-\l(p)}
\eeq
where the sum is over minimal programs $p$ for which $U$ outputs a
string starting with $x$. The so-called minimal programs are
defined similarly to the prefix programs, but $U$ need not to
halt, which is indicated by the $*$. Minimal programs are those
which are left to the input head in the moment when $U$ wrote the
last bit of $x$ \cite{Li:97,Hutter:04uaibook}.
Before we can discuss the stochastic properties of $\MM$ we
need the concept of (semi)measures for strings.

\fdefinition{defSemi}{Continuous (Semi)measures}{
$\mu(x)$ denotes the probability that a sequence starts
with string $x$. We call $\mu\geq 0$ a (continuous) semimeasure if
$\mu(\epstr)\leq 1$ and $\mu(x)\geq\sum_{a\in\X}\mu(xa)$, and a
(probability) measure if equalities hold.
}

\noindent The reason for calling $\mu$ with the above property a
probability measure is that it satisfies Kolmogorov's axioms of
probability in the following sense: The sample space is
$\X^\infty$ with elements
$\omega=\omega_1\omega_2\omega_3...\in\X^\infty$ being infinite
sequences over alphabet $\X$. The set of events (the
$\sigma$-algebra) is defined as the
set generated from the cylinder sets
$\Gamma_{x_{1:n}}:=\{\omega:\omega_{1:n}=x_{1:n}\}$ by countable
union and complement. A probability
measure $\mu$ is uniquely defined by giving its values
$\mu(\Gamma_{x_{1:n}})$ on the cylinder sets, which we abbreviate
by $\mu(x_{1:n})$. We will also call $\mu$ a measure, or even more
loose a probability distribution.

\noindent We have $\sum_{a\in\X}\MM(xa)<\MM(x)$ because there are
programs $p$ that output $x$, not followed by any $a\in\X$.
They just stop after printing $x$ or continue forever without any
further output. Together with $\MM(\epstr)=1$ this shows that $\MM$
is a semimeasure, but {\it not} a probability measure. We can now
state the fundamental property of $\MM$ \cite{Zvonkin:70,Solomonoff:78}:

\ftheorem{thUniM}{Universality of $\MM$}{
The universal prior $\MM$ is an enumerable semimeasure that
multiplicatively dominates all enumerable semimeasures in the
sense that $\MM(x) \;\geqm\; 2^{-K(\rho)}\cdot \rho(x)$
for all enumerable semimeasures $\rho$. $\MM$ is enumerable, but not
estimable (nor recursive).
}

\noindent The Kolmogorov complexity of a function like $\rho$ is
defined as the length of the shortest self-delimiting code of a
Turing machine computing this function in the sense of Definition
\ref{defCompFunc}. Up to a multiplicative constant, $\MM$ assigns higher
probability to all $x$ than any other computable probability
distribution.

It is possible to normalize $\MM$ to a true probability measure
$\MM_{norm}$ \cite{Solomonoff:78,Li:97} with dominance still being
true, but at the expense of giving up enumerability ($\MM_{norm}$
is still approximable). $\MM$ is more convenient when studying
algorithmic questions, but a true probability measure like
$\MM_{norm}$ is more convenient when studying stochastic questions.

\section{Universal Sequence Prediction}\label{secUSP}

In which sense does $\MM$ incorporate Occam's razor and Epicurus'
principle of multiple explanations? Since the shortest programs
$p$ dominate the sum in $M$, $\MM(x)$ is roughly equal to
$2^{-K(x)}$ ($\MM(x)=2^{-K(x)+O(K(\l(x))}$), i.e.\
$\MM$ assigns high probability to simple
strings. More useful is to think of $x$ as being the observed
history. We see from (\ref{Mdef}) that every program $p$
consistent with history $x$ is allowed to contribute to $\MM$
(Epicurus). On the other hand, shorter programs give significantly
larger contribution (Occam). How does all this affect prediction?
If $\MM(x)$ describes our (subjective) prior belief in $x$, then
$\MM(y|x):=\MM(xy)/\MM(x)$ must be our posterior belief in $y$.
From the symmetry of algorithmic information $K(xy)\approx
K(y|x)+K(x)$, and $\MM(x)\approx 2^{-K(x)}$ and $\MM(xy)\approx
2^{-K(xy)}$ we get $\MM(y|x)\approx 2^{-K(y|x)}$. This tells us
that $\MM$ predicts $y$ with high probability iff $y$ has an easy
explanation, given $x$ (Occam \& Epicurus).

The above qualitative discussion should not create the impression
that $\MM(x)$ and $2^{-K(x)}$ always lead to predictors of
comparable quality. Indeed, in the online/incremental setting,
$K(y)=O(1)$ invalidates the consideration above. The proof of
(\ref{eukdist}) below, for instance, depends on $\MM$ being a
semimeasure and the chain rule being exactly true, neither of them is
satisfied by $2^{-K(x)}$. See \cite{Hutter:03unimdl} for a
detailed analysis.

Sequence prediction algorithms try to predict the continuation
$x_t\in\X$ of a given sequence $x_1...x_{t-1}$.
The following bound shows that $M$ predicts computable sequences well:
\beq\label{eqDetMbnd}
  \sum_{t=1}^\infty(1\!-\!\MM(x_t|x_{<t}))^2 \;\leq\;
  -\odt \sum_{t=1}^\infty\ln \MM(x_t|x_{<t}) \;=\;
  -\odt\ln\MM(x_{1:\infty}) \;\leq\;
  \odt\ln 2\cdot \Km(x_{1:\infty}),
\eeq
where the monotone complexity
$\Km(x_{1:\infty})=\min\{\l(p):U(p)=x_{1:\infty}\}$ is defined as
the length of the shortest (nonhalting) program computing
$x_{1:\infty}$ \cite{Zvonkin:70,Levin:73random}. In the first
inequality we have used $(1-a)^2\leq-\odt\ln a$ for $0\leq a\leq
1$. In the equality we exchanged the sum with the logarithm and
eliminated the resulting product by the chain rule. In the last inequality
we used $\MM(x)\geq 2^{-\Km(x)}$, which follows from
(\ref{Mdef}) by dropping all terms in $\sum_p$ except for the
shortest $p$ computing $x$. If $x_{1:\infty}$ is a computable
sequence, then $\Km(x_{1:\infty})$ is finite, which implies
$\MM(x_t|x_{<t})\to 1$
($\sum_{t=1}^\infty(1-a_t)^2<\infty\Rightarrow a_t\to 1$). This
means, that if the environment is a computable sequence
(whichsoever, e.g.\ the digits of $\pi$ or $e$ in $\X$ary
representation), after having seen the first few digits, $\MM$
correctly predicts the next digit with high probability, i.e.\ it
recognizes the structure of the sequence.

Assume now that the true sequence is
drawn from a computable
probability distribution $\mu$, i.e.\ the true (objective)
probability of $x_{1:t}$ is $\mu(x_{1:t})$. The probability of
$x_t$ given $x_{<t}$ hence is
$\mu(x_t|x_{<t})=\mu(x_{1:t})/\mu(x_{<t})$.
Solomonoff's \cite{Solomonoff:78} central result is that $\MM$
converges to $\mu$. More precisely, for binary alphabet, he showed that
\beq\label{eukdist}
  \sum_{t=1}^\infty
  \nq\nq\;\sum_{\qquad x_{<t}\in\B^{t-1}}\nq\nq\;
  \mu(x_{<t}) \Big(\MM(0|x_{<t})-\mu(0|x_{<t})\Big)^2
  \;\leq\;
  {\odt}\ln 2\!\cdot\!K(\mu)+O(1) \;<\; \infty.
\eeq
The infinite sum can only be finite if the difference
$\MM(0|x_{<t})-\mu(0|x_{<t})$ tends to zero for $t\to\infty$ with
$\mu$-probability $1$ (see Definition~\ref{defConv}$(i)$ and
\cite{Hutter:01alpha} or Section~\ref{secConv} for general
alphabet). This holds for {\it any} computable probability
distribution $\mu$. The reason for the astonishing property of a
single (universal) function to converge to {\it any} computable
probability distribution lies in the fact that the set of
$\mu$-random sequences differ for different $\mu$. Past data
$x_{<t}$ are exploited to get a (with $t\to\infty$) improving
estimate $\MM(x_t|x_{<t})$ of $\mu(x_t|x_{<t})$.

The universality property (Theorem~\ref{thUniM}) is the central
ingredient in the proof of (\ref{eukdist}). The proof
involves the construction of a semimeasure $\xi$
whose dominance is obvious. The hard part is to show its
enumerability and equivalence to $\MM$.
Let $\M$ be the (countable) set of all enumerable semimeasures
and define
\beq\label{xidef}
  \xi(x):=\sum_{\nu\in\M}2^{-K(\nu)}\nu(x).
\eeq
Then dominance
\beq\label{xidom}
 \xi(x)\geq 2^{-K(\nu)}\nu(x)\quad\forall\,\nu\in\M
\eeq
is obvious. Is $\xi$ lower semicomputable? To answer this
question one has to be more precise. Levin \cite{Zvonkin:70} has
shown that the set of {\em all} lower semicomputable semimeasures
is enumerable (with repetitions). For this (ordered multi) set
$\M=\M_{enum}^{semi}:=\{\nu_1,\nu_2,\nu_3,...\}$ and
$K(\nu_i):=K(i)$ one can easily see that $\xi$ is lower
semicomputable. Finally proving $\MM(x)\geqm\xi(x)$ also
establishes universality of $\MM$ (see \cite{Solomonoff:78,Li:97}
for details).

The advantage of $\xi$ over $\MM$ is that it immediately
generalizes to arbitrary weighted sums of (semi)measures
for arbitrary countable $\M$.

\section{Universal (Semi)Measures}\label{secUSM}

What is so special about the set of all enumerable
semimeasures $\M_{enum}^{semi}$? The larger we choose $\M$ the less restrictive
is the assumption that $\M$ should contain the true distribution
$\mu$, which will be essential throughout the paper.
Why do not restrict to the still rather general class of estimable
or recursive (semi)measures? It is clear that for every
countable (multi)set $\M$, the universal or mixture distribution
\beq\label{defxi}
  \xi(x):=\xi_\M(x):=\sum_{\nu\in\M} w_\nu \nu(x)
  \qmbox{with} \sum_{\nu\in\M}w_\nu\leq 1 \qmbox{and} w_\nu>0
\eeq
dominates all $\nu\in\M$. This dominance is
necessary for the desired convergence $\xi\to\mu$ similarly to
(\ref{eukdist}). The question is what properties $\xi$ possesses.
The distinguishing property of $\M_{enum}^{semi}$ is that $\xi$ is
itself an element of $\M_{enum}^{semi}$. When concerned with
predictions, $\xi_\M\in\M$ is not by itself an important property,
but whether $\xi$ is computable in one of the senses of Definition
\ref{defCompFunc}. We define
\bqan
 \M_1\geqm\M_2 & :\Leftrightarrow &
 \mbox{there is an element of $\M_1$ that dominates all elements of
 $\M_2$} \\
 & :\Leftrightarrow &
\exists\rho\!\in\!\M_1\;\forall\nu\!\in\!\M_2\;\exists w_\nu\!>\!0
\;\forall x:\rho(x)\!\geq\!w_\nu\nu(x).
\eqan
$\geqm $ is transitive (but not necessarily reflexive) in the
sense that $\M_1 \geqm \M_2 \geqm \M_3$ implies $\M_1 \geqm \M_3$
and $\M_0 \supseteq \M_1 \geqm \M_2 \supseteq \M_3$ implies $\M_0
\geqm \M_3$.
For the computability concepts introduced in Section~\ref{secCC}
we have the following proper set inclusions
\beqn
\begin{array}{ccccccc}
  \M_{rec}^{msr}  & \subset & \M_{est}^{msr}  & \equiv  & \M_{enum}^{msr}  & \subset & \M_{appr}^{msr} \\
        \cap       &         &      \cap       &         &       \cap       &         &     \cap        \\
  \M_{rec}^{semi} & \subset & \M_{est}^{semi} & \subset & \M_{enum}^{semi} & \subset & \M_{appr}^{semi}
\end{array}
\eeqn
where $\M^{msr}_c$ stands for the set of all probability measures
of appropriate computability type $c\in\{$rec=recursive, est=estimable, enum=enumerable,
appr=approximable$\}$, and similarly for semimeasures
$\M^{semi}_c$. From an enumeration of a measure $\rho$ one can
construct a co-enumeration by exploiting
$\rho(x_{1:n})=1-\sum_{y_{1:n}\neq x_{1:n}}\rho(y_{1:n})$. This
shows that every enumerable measure is also co-enumerable, hence
estimable, which proves the identity $\equiv$ above.

With this notation, Theorem~\ref{thUniM} implies
$\M_{enum}^{semi}\geqm\M_{enum}^{semi}$. Transitivity allows to
conclude, for instance, that
$\M_{appr}^{semi}\geqm\M_{rec}^{msr}$, i.e.\ that there is an
approximable semimeasure that dominates all recursive measures.

The standard ``diagonalization'' way of proving
$\M_1\ngeqm\M_2$ is to take an arbitrary
$\mu\in\M_1$ and ``increase'' it to $\rho$ such that
$\mu\ngeqm\rho$ and show that $\rho\in\M_2$.
There are $7\times 7$ combinations of (semi)measures $\M_1$ with
$\M_2$ for which $\M_1\geqm\M_2$ could be true or false. There are
four basic cases, explicated in the following theorem, from which
the other 49 combinations displayed in Table~\ref{tabUniSMsr}
follow by transitivity.

\ftheorem{thNoUniApp}{Universal (semi)measures}{
A semimeasure $\rho$ is said to be universal for $\M$ if it
multiplicatively dominates all elements of $\M$ in the sense
$\forall\nu\exists w_\nu>0:\rho(x)\geq w_\nu\nu(x)\forall x$. The
following holds true:
\begin{list}{}{\parsep=1ex}
\item[$o)$]
$\exists\rho:\{\rho\}\geqm\M$: For every countable set
of (semi)measures $\M$, there is a (semi)measure that dominates
all elements of $\M$.
\item[$i)$]
$\M_{enum}^{semi}\geqm\M_{enum}^{semi}$:
The class of enumerable semimeasures {\em contains}
a universal element.
\item[$ii)$]
$\M_{appr}^{msr}\geqm\M_{enum}^{semi}$:
There {\em is} an approximable measure that dominates all enumerable
semimeasures.
\item[$iii)$]
$\M_{est}^{semi}\ngeqm\M_{rec}^{msr}$: There is
{\em no} estimable semimeasure that dominates all recursive
measures.
\item[$iv)$]
$\M_{appr}^{semi}\ngeqm\M_{appr}^{msr}$: There is
{\em no} approximable semimeasure that dominates all approximable
measures.
\end{list}
}

\begin{table}[thb]
\ftablex{tabUniSMsr}{Existence of universal (semi)measures}{%
The entry in row $r$ and column $c$ indicates whether there is an
$r$-able (semi)measure $\rho$ dominating the set $\M$ that contains all
$c$-able (semi)measures, where $r,c\in\{$recurs, estimat, enumer,
approxim$\}$. Enumerable measures are estimable. This is the
reason why the enum.\ row and column in case of measures are
missing. The superscript indicates from which part of Theorem
\ref{thNoUniApp} the answer follows. For the bold face entries
directly, for the others using transitivity of $\geqm $.
\begin{center}
\begin{tabular}{|c|c||c|c|c|c||c|c|c|}\hline
      $\nwarrow$ &  $\M$ & \multicolumn{4}{c||}{semimeasure} & \multicolumn{3}{c|}{measure}\\ \hline
$\rho$&$\searrow$& rec.      & est.       & enum.         & appr.     & rec.          & est.       & appr.        \\ \hline\hline
      s  & rec. & no$^{iii}$ & no$^{iii}$ & no$^{iii}$    & no$^{iv}$ & no$^{iii}$    & no$^{iii}$ & no$^{iv}$    \\ \cline{2-9}
      e  & est.  & no$^{iii}$ & no$^{iii}$ & no$^{iii}$    & no$^{iv}$ & {\bf no}$^{\bf iii}$& no$^{iii}$ & no$^{iv}$    \\ \cline{2-9}
      m  & enum. & yes$^{i}$  & yes$^{i}$  & {\bf yes}$^{\bf i}$ & no$^{iv}$ & yes$^{i}$     & yes$^{i}$  & no$^{iv}$    \\ \cline{2-9}
      i  &appr.  & yes$^{i}$  & yes$^{i}$  & yes$^{i}$     & no$^{iv}$ & yes$^{i}$     & yes$^{i}$  & {\bf no}$^{\bf iv}$\\ \hline\hline
      m  & rec.  & no$^{iii}$ & no$^{iii}$ & no$^{iii}$    & no$^{iv}$ & no$^{iii}$    & no$^{iii}$ & no$^{iv}$    \\ \cline{2-9}
      s  & est.  & no$^{iii}$ & no$^{iii}$ & no$^{iii}$    & no$^{iv}$ & no$^{iii}$    & no$^{iii}$ & no$^{iv}$    \\ \cline{2-9}
      r  &appr.  & yes$^{ii}$ & yes$^{ii}$ & {\bf yes}$^{\bf ii}$& no$^{iv}$ & yes$^{ii}$    & yes$^{ii}$ & no$^{iv}$    \\ \hline
\end{tabular}
\end{center}}
\end{table}

\noindent If we ask for a universal (semi)measure that at least
satisfies the weakest form of computability, namely being
approximable, we see that the largest dominated set among the 7
sets defined above is the set of enumerable semimeasures. This is
the reason why $\M_{enum}^{semi}$ plays a special role. On the
other hand, $\M_{enum}^{semi}$ is not the largest set dominated by
an approximable semimeasure, and indeed no such largest set
exists. One may, hence, ask for ``natural'' larger sets $\M$. One
such set, namely the set of cumulatively enumerable semimeasures
$\M_{\text{CEM}}$, has recently been discovered by Schmidhuber
\cite{Schmidhuber:00toe,Schmidhuber:02gtm}, for which even
$\xi_{\text{CEM}}\in\M_{\text{CEM}}$ holds.

\noindent Theorem~\ref{thNoUniApp} also holds for {\em discrete
(semi)measures} $P$ defined as follows:

\fdefinition{defDSemi}{Discrete (semi)measures}{
$P(x)$ denotes the probability of $x\in\SetN$. We call
$P:\SetN\to[0,1]$ a discrete (semi)measure if $\sum_{x\in\SetN}
P(x)\stackrel{(<)}=1$.
}
Theorem~\ref{thNoUniApp}
$(i)$ is Levin's major result \cite[Thm.4.3.1 \& Thm.4.5.1]{Li:97}, %
and $(ii)$ is due to Solomonoff \cite{Solomonoff:78}. %
The proof of $\M_{rec}^{semi}\ngeqm\M_{rec}^{semi}$ in
\cite[p249]{Li:97} contains minor errors and is not extensible to
$(iii)$, and the proof in \cite[p276]{Li:97} only applies to
infinite alphabet and not to the binary/finite case considered
here. $\M_{est}^{semi}\ngeqm\M_{est}^{semi}$
is mentioned in \cite{Zvonkin:70} without proof.
A direct proof of $(iv)$ can be found in \cite{Hutter:04uaibook}.
Here, we reduce $(iv)$ to $(iii)$ by exploiting the following
elementary fact (well-known for integer-valued functions, see
e.g.\ \cite[p634]{Simpson:77}):

\flemma{lemOracle}{Approximable = $H$-estimable}{
A function is approximable iff it is estimable with the help of
the halting oracle.
}

\paradot{Proof}
With $H$-computable we mean, computable with the help of the
halting oracle, or equivalently, computable under extra input of
the halting sequence $h=h_{1:\infty}\in\B^\infty$, where $h_n=1$
$:\Leftrightarrow$ $U(n)$ halts.

Assume $f$ is approximable, i.e.\ $\forall\eps\exists y,m:
R(m,y,\eps)$, where relation $R(m,y,\eps):=[\forall n\geq
m:|f_n(x)-y|<\eps]$ and recursive $f_n\to f$. Fix $\eps>0$.
Search (dovetail) for $m\in\SetN$ and $y$ ($\in\odt\eps\SetZ$ is
sufficient) such that $R(m,y,\eps)=$true. $R$ is
co-enumerable, hence $H$-decidable, hence $y$ can be $H$-computed,
hence $f$ is $H$-estimable, since $f(x)=y\pm O(\eps)$.

Now assume that $f$ is $H$-estimable, i.e.\ $\exists T\in$TM
$\forall\eps,x:|T(x,\eps,h)-f(x)|<\eps$. Since $h$ is
co-enumerable, $T$ and hence $f$ are approximable. More formally,
let $h_n^t=1$ $:\Leftrightarrow$ $U(n)$ halts within $t$ steps.
Then $g(x,\eps) := T(x,\eps,h) = T(x,\eps,\lim_{t\to\infty}h^t) =
\lim_{t\to\infty}T(x,\eps,h^t)$ is approximable, where the
exchange of limits holds, since $T$ only reads $n_{x\eps}<\infty$
bits of $h$ and $h_{1:n_{x\eps}}=h^t_{1:n_{x\eps}}$ for
sufficiently large $t$. \qed

\section{Proof of Theorem~\ref{thNoUniApp}}\label{secProof}

We first prove the theorem for discrete (semi)measures $P$ (Definition
\ref{defDSemi}), since it contains the essential ideas in a
cleaner form. We then present the proof for continuous
(semi)measures $\mu$ (Definition~\ref{defSemi}). We present proofs
for binary alphabet $\X=\B$ only. The proofs naturally generalize from
binary to arbitrary finite alphabet. $\arg\min_x f(x)$ is the $x$
that minimizes $f(x)$. Ties are broken in an arbitrary but
computable way (e.g.\ by taking the smallest $x$).

\paradot{Proof (discrete case)}\\%
\paranodot{(o)} $Q(x):=\sum_{P\in\M}w_P P(x)$
with $w_P>0$ obviously dominates all $P\in\M$ (with constant
$w_P$). With $\sum_P w_P=1$ and all $P$ being discrete
(semi)measures also $Q$ is a discrete (semi)measure.

\paranodot{(i)} See \cite[Thm.4.3.1]{Li:97}.

\paranodot{(ii)} Let $P$ be the universal element in
$\M_{enum}^{semi}$ and $\alpha:=\sum_x P(x)$. We normalize $P$ by
$Q(x):={1\over\alpha}P(x)$. Since $\alpha\leq 1$ we have $Q(x)\geq
P(x)$. Hence $Q\geq P\geqm\M_{enum}^{semi}$. As a
ratio between two enumerable functions, $Q$ is still approximable,
hence $\M_{appr}^{msr}\geqm\M_{enum}^{semi}$.

\paranodot{(iii)}
Let $P\in\M_{rec}^{semi}$. We partition $\SetN$ into chunks
$I_n:=\{2^{n-1},...,2^n-1\}$ ($n\geq 1$) of increasing size. With
$x_n:=\arg\min_{x\in I_n}P(x)$ we define $Q(x_n):={1\over
n(n+1)}\forall n$ and $Q(x):=0$ for all other $x$. Exploiting that
a minimum is smaller than an average and that $\mu$ is a
semimeasure, we get
\beqn
P(x_n)=\min_{x\in I_n}P(x)\leq{1\over|I_n|}\sum_{x\in
I_n}P(x)\leq{1\over|I_n|}={1\over 2^{n-1}}= {n(n+1)\over
2^{n-1}} Q(x_n)
\eeqn
Since ${n(n+1)\over 2^{n-1}}\to 0$ for $n\to\infty$, $P$ cannot
dominate $Q$ ($P\ngeqm Q$). With $P$ also $Q$
is recursive. Since $P$ was an arbitrary recursive semimeasure
and $Q$ is a recursive measure ($\sum Q(x)=\sum[{1\over
n(n+1)}]=\sum[{1\over n}-{1\over n+1}]=1$) this implies
$\M_{rec}^{semi}\ngeqm\M_{rec}^{msr}$.

Assume now that there is an estimable semimeasure
$S\geqm\M_{rec}^{msr}$. We construct a recursive semimeasure
$P\geqm S$ as follows. Choose an initial $\eps>0$ and finitely
compute an $\eps$-approximation $\hat S$ of $S(x)$. If $\hat
S>2\eps$ define $P(x):=\odt\hat S$, else halve $\eps$ and repeat
the process. Since $S(x)>0$ (otherwise it could not dominate,
e.g.\ $T(x):={1\over x(x+1)}\in\M_{rec}^{msr}$) the loop
terminates after finite time. So $P$ is recursive. Inserting $\hat
S=2P(x)$ and $\eps<\odt\hat S=P(x)$ into $|S(x)-\hat S|<\eps$ we
get $|S(x)-2P(x)|<P(x)$, which implies $S(x)\geq P(x)$ and
$S(x)\leq 3P(x)$. The former implies $\sum_x P(x)\leq \sum_x
S(x)\leq 1$, i.e.\ $P$ is a semimeasure. The latter implies
$P\geq{1\over 3}S\geqm\M_{rec}^{msr}$. Hence $P$ is a recursive
semimeasure dominating all recursive measures, which contradicts
what we have proven in the first half of $(iii)$. Hence the
assumption on $S$ was wrong which establishes
$\M_{est}^{semi}\ngeqm\M_{rec}^{msr}$.

\paranodot{(iv)} From $(iii)$ we know that
$\M_{est}^{semi}\ngeqm\M_{est}^{msr}$. The proof and hence result
remains valid under the halting oracle, i.e.\
$\M_{H\text{-}est}^{semi}\ngeqm\M_{H\text{-}est}^{msr}$. By Lemma
\ref{lemOracle}, the $H$-estimable functions/(semi)measures coincide
with the approximable functions/(semi)measures, hence
$\M_{appr}^{semi}\ngeqm\M_{appr}^{msr}$. \qed

\paradot{Proof (continuous case)}\\%
The major difference to the discrete case is that one also has to
take care that $\rho(x)\stackrel{(>)}=\rho(x0)+\rho(x1)$, $x\in\B^*$, is
respected. On the other hand, the chunking $I_n:=\B^n$ is more
natural here.

\paranodot{(o)} $\rho(x):=\sum_{\nu\in\M}w_\nu \nu(x)$ with $w_\nu>0$
obviously dominates all $\nu\in\M$ (with domination constant
$w_\nu$). With $\sum_\nu w_\nu=1$ and all $\nu$ being
(semi)measures also $\rho$ is a (semi)measure.

\paranodot{(i)} See \cite[Thm.4.5.1]{Li:97}.

\paranodot{(ii)} Let $\xi$ be a universal element in $\M_{enum}^{semi}$.
We define \cite{Solomonoff:78}
\beqn
  \xi_{norm}(x_{1:n}) \;:=\;
  \prod_{t=1}^n{\xi(x_{1:t}) \over \xi(x_{<t}0)+\xi(x_{<t}1)}.
\eeqn
By induction one can show that $\xi_{norm}$ is a measure and
that $\xi_{norm}(x)\geq\xi(x)\forall x$, hence
$\xi_{norm}\geq\xi\geqm\M_{enum}^{semi}$. As a ratio
of enumerable functions, $\xi_{norm}$ is still approximable, hence
$\M_{appr}^{msr}\geqm\M_{enum}^{semi}$.

\paranodot{(iii)} Analogous to the discrete case we could start by
recursively defining $x_k^*:=\arg\min_{x_k}\mu(x_{<k}^*x_k)$ for
$\mu\in\M_{rec}^{semi}$. See \cite{Hutter:03unipriors} for a proof
along this line.
Simpler is to directly consider $\mu\in\M_{est}^{semi}$ and to
compute $x^*_{1:\infty}$ recursively by computing some
$\eps$-approximation $e(x_k|x^*_{<t})$ of $\mu(x_k|x^*_{<t})$ and
define $x^*_k=\arg\max_{x_k}e(x_k|x^*_{<t})$, which implies
$\mu(x^*_k|x^*_{<t})\leq\odt+\eps$. Finally we define measure
$\rho$ by $\rho(x_{1:k}^*)=1\forall k$ and $\rho(x)=0$ for all $x$
that are not prefixes of $x_{1:\infty}^*$.
Hence
$\mu(x_{1:n}^*)\leq(\odt+\eps)^n=(\odt+\eps)^n\rho(x_{1:n}^*)$,
which demonstrates that $\mu$ does not dominate $\rho$ for
$\eps<\odt$. Since $\mu\in\M_{est}^{semi}$ was arbitrary and
$\rho$ is a recursive measure, this implies
$\M_{est}^{semi}\ngeqm\M_{rec}^{msr}$.

\paranodot{(iv)} Identical to discrete case. \qed

\section{Posterior Convergence}\label{secConv}

We investigated in detail the computational properties of
various mixture distributions $\xi$. A mixture $\xi_\M$
multiplicatively dominates all distributions in $\M$. We
mentioned that dominance implies posterior convergence. In this
section we present in more detail what dominance implies and what
not.

Convergence of $\xi(x_t|x_{<t})$ to $\mu(x_t|x_{<t})$ with
$\mu$-probability 1 tells us that $\xi(x_t|x_{<t})$ is close to
$\mu(x_t|x_{<t})$ for sufficiently large $t$ on `most'
sequences $x_{1:\infty}$. It says nothing about the speed of
convergence, nor whether convergence is true for any {\em particular}
sequence (of measure 0). Convergence {\em in mean sum} defined
below is intended to capture the rate of convergence,
Martin-L\"{o}f randomness is used to capture convergence
properties for individual sequences.

Martin-L\"{o}f randomness is a very important concept of
randomness of individual sequences, which is closely related to
Kolmogorov complexity and Solomonoff's universal prior. Levin gave
a characterization equivalent to Martin-L\"{o}f's original
definition \cite{Levin:73random}:

\ftheorem{defML}{Martin-L\"{o}f random sequences}{
A sequence $x_{1:\infty}$ is $\mu$-Martin-L\"{o}f random
($\mu$.M.L.) iff there is a constant $c$ such that
$\MM(x_{1:n})\leq c\cdot \mu(x_{1:n})$ for all $n$.
}

\noindent An  equivalent formulation for estimable $\mu$ is:
\beq\label{KmMLr}
  x_{1:\infty} \mbox{ is $\mu$.M.L.-random}
  \quad\Leftrightarrow\quad
  \Km(x_{1:n})= -\log\mu(x_{1:n})+O(1) \;\forall n
\eeq
Theorem~\ref{defML} follows from
(\ref{KmMLr}) by exponentiation, ``using $2^{-\Km}\approx\MM$''
and noting that $\MM\geqm\mu$ follows from universality of $\MM$.
Consider the special case of $\mu$ being a fair coin, i.e.\
$\mu(x_{1:n})=2^{-n}$, then $x_{1:\infty}$ is M.L.\ random {\em
iff} $\Km(x_{1:n})=n+O(1)$, i.e.\ if $x_{1:n}$ is incompressible.
For general $\mu$, $-\lb\mu(x_{1:n})$ is the length of the
Shannon-Fano code of $x_{1:n}$, hence $x_{1:\infty}$ is
$\mu$.M.L.-random {\em iff} the Shannon-Fano code is optimal.

One can show that a $\mu$.M.L.-random sequence $x_{1:\infty}$
passes {\em all} thinkable effective randomness tests, e.g.\ the
law of large numbers, the law of the iterated logarithm, etc.
In particular, the set of all $\mu$.M.L.-random sequences has
$\mu$-measure 1.
The following generalization is natural when considering general
Bayes mixtures $\xi$ as in this work:

\fdefinition{defmuMr}{$\mu/\xi$-random sequences}{
A sequence $x_{1:\infty}$ is called $\mu/\xi$-random
($\mu.\xi$.r.) iff there is a constant $c$ such that
$\xi(x_{1:n})\leq c\cdot \mu(x_{1:n})$ for all $n$.
}

Typically, $\xi$ is a mixture over some $\M$ as defined in
(\ref{defxi}), in which case the reverse inequality
$\xi(x)\geqm\mu(x)$ is also true (for all $x$). For finite $\M$ or
if $\xi\in\M$, the definition of $\mu/\xi$-randomness depends only
on $\M$, and not on the specific weights $w_\nu$ used in $\xi$. For
$\M=\M_{enum}^{semi}$, $\mu/\xi$-randomness is just
$\mu$.M.L.-randomness. The larger $\M$, the more patterns are
recognized as nonrandom.
Roughly speaking, those regularities characterized by some
$\nu\in\M$ are recognized by $\mu/\xi$-randomness, i.e.\ for
$\M\subset\M_{enum}^{semi}$ some $\mu/\xi$-random strings may not
be M.L.\ random.
Other randomness concepts, e.g.\ those by Schnorr, Ko, van
Lambalgen, Lutz, Kurtz, von Mises, Wald, and Church (see
\cite{Wang:96,Lambalgen:87,Schnorr:71}), could possibly also be
characterized in terms of $\mu/\xi$-randomness for particular
choices of $\cal M$.

A classical (nonrandom)
real-valued sequence $a_t$ is defined to converge to $a_*$, short
$a_t\to a_*$ if $\forall\eps\exists t_0\forall t\geq
t_0:|a_t-a_*|<\eps$. We are interested in convergence properties
of random sequences $z_t(\omega)$ for $t\to\infty$ (e.g.\
$z_t(\omega)=\xi(\omega_t|\omega_{<t})-\mu(\omega_t|\omega_{<t})$).
We denote $\mu$-expectations by $\E$. The expected value of a
function $f:\X^t\to\SetR$, dependent on $x_{1:t}$, independent of
$x_{t+1:\infty}$, and possibly undefined on a set of $\mu$-measure
0, is $\E[f] =
\sumprime_{\!x_{1:t}\in\X^t}\mu(x_{1:t})f(x_{1:t})$. The prime
denotes that the sum is restricted to $x_{1:t}$ with
$\mu(x_{1:t})\neq 0$. Similarly we use $\P[..]$ to denote the
$\mu$-probability of event $[..]$.
We define four convergence concepts for random sequences.

\fdefinition{defConv}{Convergence of random sequences}{
Let $z_1(\omega),z_2(\omega),...$ be a sequence of real-valued
random variables. $z_t$ is said to
converge for $t\to\infty$ to (random variable) $z_*$
\begin{list}{}{\itemsep=1ex\leftmargin=8ex}
\item[$i)$] with probability 1 (w.p.1) $:\Leftrightarrow$
  $\P[\{\omega:z_t\to z_*\}]=1$,
\item[$ii)$] in mean sum (i.m.s.) $:\Leftrightarrow$
$\sum_{t=1}^\infty\E[(z_t-z_*)^2]<\infty$,
\item[$iii)$] for every $\mu$-Martin-L{\"o}f random sequence ($\mu$.M.L.) $:\Leftrightarrow$ \\
$\forall\omega:$ If $[\exists c\forall n:
\MM(\omega_{1:n})\leq c\mu(\omega_{1:n})]$
  then $z_t(\omega)\to z_*(\omega)$ for $t\to\infty$,
\item[$iv)$] for every $\mu/\xi$-random sequence ($\mu.\xi$.r.) $:\Leftrightarrow$ \\
$\forall\omega:$ If $[\exists c\forall n:
\xi(\omega_{1:n})\leq c\mu(\omega_{1:n})]$
  then $z_t(\omega)\to z_*(\omega)$ for $t\to\infty$.
\end{list}
}

\noindent In statistics, $(i)$ is the ``default'' characterization of
convergence of random sequences.
Convergence i.m.s.\ $(ii)$ is very strong: it provides a rate of
convergence in the sense that the expected number of times $t$ in
which $z_t$ deviates more than $\eps$ from $z_*$ is finite and
bounded by $c/\eps^2$ and the probability that the number of
$\eps$-deviations exceeds $c\over\eps^2\delta$ is smaller than
$\delta$, where $c:=\sum_{t=1}^\infty\E[(z_t-z_*)^2]$.
Nothing can be said for {\em which} $t$ these deviations occur.
If, additionally, $|z_t-z_*|$ were monotone decreasing, then
$|z_t-z_*|=o(t^{-1/2})$ could be concluded.
$(iii)$ uses Martin-L\"{o}f's notion of randomness of {\em individual}
sequences to define convergence M.L. Since this work
deals with general Bayes mixtures $\xi$, we generalized in $(iv)$
the definition of convergence M.L.\ based on $\MM$ to
convergence $\mu.\xi$.r.\ based on $\xi$ in a natural way.
One can show that convergence i.m.s.\ implies convergence w.p.1.
Also convergence M.L.\ implies convergence w.p.1.
Universality of $\xi$ implies the following posterior convergence results:


\ftheorem{thConv}{Convergence of $\xi$ to $\mu$}{
Let there be sequences $x_1x_2...$ over a finite alphabet $\X$
drawn with probability $\mu(x_{1:n})\in\M$ for the first $n$
symbols, where $\mu$ is a measure and $\M$ a countable set of
(semi)measures. The universal/mixture posterior probability
$\xi(x_t|x_{<t})$
of the next symbol $x_t$ given $x_{<t}$
is related to the true posterior probability $\mu(x_t|x_{<t})$
in the following way:\vspace{-1ex}
\beqn
   \sum_{t=1}^n\E{\textstyle\left[\left(\sqrt{{\xi(x_t|x_{<t})
          \over\mu(x_t|x_{<t})}}-1\right)^2\right]} \;\leq\;
   \sum_{t=1}^n\E\bigg[\sum_{x'_t}
        \left(\sqrt{\xi(x'_t|x_{<t})}-\sqrt{\mu(x'_t|x_{<t})}\right)^2\bigg]
        \;\leq\; \ln{w_\mu^{-1}} \;<\; \infty
\eeqn
where $w_\mu$ is the weight (\ref{defxi}) of $\mu$ in $\xi$.
}

\noindent Theorem~\ref{thConv} implies
\beqn
 \mbox{$\sqrt{\xi(x'_t|x_{<t})} \to \sqrt{\mu(x'_t|x_{<t})}$
 for any $x'_t$ and
 $\sqrt{{\xi(x_t|x_{<t})\over\mu(x_t|x_{<t})}} \to 1$, both
 i.m.s.\ for $t\to\infty$}.
\eeqn
\noindent The latter strengthens the result
$\xi(x_t|x_{<t})/\mu(x_t|x_{<t})\to 1$ w.p.1 derived by G\'acs
\cite[Thm.5.2.2]{Li:97} in that it also provides the ``speed'' of
convergence.

Note also the subtle difference between the two convergence
results. For {\em any} sequence $x'_{1:\infty}$ (possibly constant
and not necessarily $\mu$-random),
$\mu(x'_t|x_{<t})-\xi(x'_t|x_{<t})$ converges to zero w.p.1
(referring to $x_{1:\infty}$), but no statement is possible for
$\xi(x'_t|x_{<t})/\mu(x'_t|x_{<t})$, since
$\lim\,\inf\mu(x'_t|x_{<t})$ could be zero. On the other hand, if
we stay {\em on}-sequence ($x'_{1:\infty} =
x_{1:\infty}$), we have $\xi(x_t|x_{<t})/\mu(x_t|x_{<t})
\to 1$ w.p.1 (whether $\inf\mu(x_t|x_{<t})$ tends to zero or not does
not matter).
Indeed, it is easy to give an example where
$\xi(x'_t|x_{<t})/\mu(x'_t|x_{<t})$ diverges. If we choose
\beqn
  \M=\{\mu_1,\mu_2\},\quad
  \mu\!\equiv\!\mu_1,\quad
  \mu_1(1|x_{<t})=\odt t^{-3} \qmbox{and}
  \mu_2(1|x_{<t})=\odt t^{-2}
\eeqn
the contribution of $\mu_2$ to $\xi$ causes $\xi$ to fall
off like $\mu_2 \sim t^{-2}$, much slower than $\mu \sim
t^{-3}$ causing the quotient to diverge:
\bqan
\mu_1(0_{1:n}) &\!=\!& \prod_{t=1}^n(1-\odt
t^{-3})\stackrel{n\to\infty}\longrightarrow c_1=0.450...>0
\;\Rightarrow\; 0_{1:\infty}\;\mbox{is a
$\mu$-random sequence},
\\
\mu_2(0_{1:n}) &\!=\!& \prod_{t=1}^n(1\!-\!\odt
t^{-2})\stackrel{n\to\infty}\longrightarrow c_2=0.358...>0
\;\Rightarrow\; \xi(0_{1:n})
\to w_1c_1+w_2c_2=:c_\xi>0
\\
\xi(0_{<t}1) &\!=\!&
w_1\mu_1(1|0_{<t})\mu_1(0_{<t})+w_2\mu_2(1|0_{<t})\mu_2(0_{<t})\to
\odt w_2c_2 t^{-2}
\eqan
\beqn
\Rightarrow \quad\xi(1|0_{<t})= {\xi(0_{<t}1)\over \xi(0_{<t})}
\rightarrow {w_2c_2\over 2c_\xi}t^{-2}
\quad\Rightarrow\quad
{\xi(1|0_{<t})\over\mu(1|0_{<t})}\to {w_2c_2\over c_\xi}t\to\infty\quad \mbox{diverges}.
\eeqn

\paradot{Proof}
For a probability distribution $y_i\geq 0$ with $\sum_i y_i=1$ and a
semi-distribution $z_i\geq 0$ with $\sum_i z_i\leq 1$ and
$i=\{1,...,N\}$, the Hellinger distance $h(\vec
y,\vec z):=\sum_i(\sqrt{y_i}-\sqrt{z_i})^2$ is upper bounded by the relative
entropy $d(\vec
y,\vec z)=\sum_i y_i\ln{y_i\over z_i}$ (and $0\ln{0\over z}:=0$).
This can be seen as follows: For arbitrary $0\leq y\leq 1$ and
$0\leq z\leq 1$ we define
\bqan
  f(y,z) &:=& y\ln{y\over z}-(\sqrt{y}-\sqrt{z})^2+z-y =
  2y g(\sqrt{z/y})
\\
  \qmbox{with}
  g(t) &:=& -\ln t+t-1\geq 0.
\eqan
This shows $f\geq 0$,
and hence $\sum_i f(y_i,z_i)\geq 0$, which implies
\beqn
  \sum_i y_i\ln{y_i\over z_i}-\sum_i(\sqrt{y_i}-\sqrt{z_i})^2 \geq
  \sum_i y_i- \sum_i z_i \geq 1-1 = 0.
\eeqn
The (conditional) $\mu$-expectations of a function $f:\X^t\to\SetR$ are defined as
\beqn
 \E[f]=\sumprime_{x_{1:t}\in\X^t}\!\!\mu(x_{1:t})f(x_{1:t})
 \qmbox{and}
  \E_t[f]:=\E[f|x_{<t}]=\sumprime_{x_t\in\X}\mu(x_t|x_{<t})f(x_{1:t}),
\eeqn
where $\sumprime$ sums over all $x_t$ or $x_{1:t}$ for which
$\mu(x_{1:t})\neq 0$.
If we insert
$\X=\{1,...,N\}$,
$N=|\X|$,
$i=x_t$,
$y_i=\mu_t:=\mu(x_t|x_{<t})$, and
$z_i=\xi_t:=\xi(x_t|x_{<t})$
into $h$ and $d$ we get (w.p.1)
\beqn\label{distdD}
  h_t(x_{<t}) \;:=\; \textstyle \sum_{x_t}
  (\sqrt{\mu_t}\!-\!\sqrt{\xi_t})^2 \qquad \leq \qquad
  d_t(x_{<t}) \;:=\; \textstyle
  \sum_{x_t}\mu_t\ln{\mu_t \over \xi_t} =
  \E_t[\ln{\mu_t\over\xi_t}].
\eeqn
Taking the expectation $\E$ and the sum $\sum_{t=1}^n$ we get
\beq\label{entropyapp}
  \sum_{t=1}^n
  \E[d_t(x_{<t})] =
  \sum_{t=1}^n\E[\E_t[
  \ln{\mu_t\over\xi_t}]] =
  \E[
  \ln \prod_{t=1}^n{\mu_t\over\xi_t}] =
  \E[
  \ln{\mu(x_{1:n}) \over \xi(x_{1:n})}] \leq
  \ln{w_\mu^{-1}}
\eeq
where we have used $\E[\E_t[..]]=\E[..]$ and exchanged the $t$-sum
with the expectation $\E$, which transforms to a product inside
the logarithm. In the last equality we have used the chain rule for
$\mu$ and $\xi$. Using universality $\xi(x_{1:n})\geq
w_\mu\mu(x_{1:n})$ yields the final inequality. Finally
\beqn
  \E_t\bigg[\Big(\sqrt{\xi_t\over \mu_t}-1\Big)^2\bigg] =
  \sum_{x_t}\!'\mu_t
  \Big(\sqrt{\xi_t\over \mu_t}-1\Big)^2  =
  \sum_{x_t}\!'(\sqrt{\xi_t}-\sqrt{\mu_t})^2 \leq
  h_t(x_{<t})\leq
  d_t(x_{<t}).
\eeqn
Taking the expectation $\E$ and the sum $\sum_{t=1}^n$ and
chaining the result with (\ref{entropyapp}) yields Theorem
\ref{thConv}. \qed

\section{Convergence in Martin-L{\"o}f Sense}\label{secMLconv}

An interesting open question is whether $\xi$ converges to $\mu$
(in difference or ratio) individually for all Martin-L\"{o}f
random sequences. Clearly, convergence $\mu$.M.L. may at most fail
for a set of sequences with $\mu$-measure zero. A convergence
M.L.\ result would be particularly interesting and natural for
Solomonoff's universal prior $M$, since M.L.\ randomness can be
defined in terms of $\MM$ (see Theorem~\ref{defML}). Attempts to
convert the bounds in Theorem~\ref{thConv} to effective
$\mu$.M.L.-randomness tests fail, since $M(x_t|x_{<t})$ is not
enumerable. The proof of $M/\mu\stackrel{M.L.}\longrightarrow 1$
given in \cite[Thm.5.2.2]{Li:97} and \cite[Thm.10]{Vitanyi:00} is
incomplete.$\!$\footnote{The formulation of their theorem is quite
misleading in general: ``{\it Let $\mu$ be a positive recursive
measure. If the length of $y$ is fixed and the length of $x$ grows
to infinity, then $M(y|x)/\mu(y|x)\to 1$ with $\mu$-probability
one. The infinite sequences $\omega$ with prefixes $x$ satisfying
the displayed asymptotics are precisely [`$\Rightarrow$' {\em and}
`$\Leftarrow$'] the $\mu$-random sequences.}'' First, for
off-sequence $y$ convergence w.p.1 does not hold ($xy$ must be
demanded to be a prefix of $\omega$). Second, the proof of
`$\Leftarrow$' has gaps (see main text). Last, `$\Rightarrow$' is
given without proof and is wrong \cite{Hutter:04mlconvx}. Also the assertion
in \cite[Thm.5.2.1]{Li:97} that $S_t:=\E\sum_{x'_t}
(\mu(x'_t|x_{<t})-M(x'_t|x_{<t}))^2$ converges to zero faster than
$1/t$ cannot be made, since $S_t$ does not decrease
monotonically \cite[Prob.2.7]{Hutter:04uaibook}. For example, for
$a_t:=1/\sqrt{t}$ if $t$ is a cube and 0 otherwise, we have
$\sum_{t=1}^\infty a_t<\infty$, but $a_t\neq o(1/t)$.} The
implication ``$\MM(x_{1:n})\leq c\cdot\mu(x_{1:n})\forall
n\Rightarrow \lim_{n\to\infty}\MM(x_{1:n})/\mu(x_{1:n})$ exists''
has been used, but not proven, and is indeed generally
wrong \cite{Hutter:04mlconvx}.
Theorem~\ref{defML} only implies
$\sup_n\MM(x_{1:n})/\mu(x_{1:n})<\infty$ for M.L.\ random
sequences $x_{1:\infty}$, and \cite[pp.\ 324--325]{Doob:53}
implies only that $\lim_{n\to\infty}\MM(x_{1:n})/\mu(x_{1:n})$ exists
w.p.1, and not $\mu$.M.L.
Vovk \cite{Vovk:87} shows that for two estimable
semimeasures $\mu$ and $\rho$ and $x_{1:\infty}$ being $\mu$
{\em and} $\rho$ M.L.\ random that
\beqn
\sum_{t=1}^\infty\sum_{x'_t}\left(\sqrt{\mu(x'_t|x_{<t})}-\sqrt{\rho(x'_t|x_{<t})}\right)^2<\infty
\qmbox{and}
\sum_{t=1}^\infty\left({\rho(x_t|x_{<t})\over\mu(x_t|x_{<t})}-1\right)^2<\infty.
\eeqn
If $\MM$ were estimable, then this would imply posterior
$\MM\to\mu$ and $\MM/\mu\to 1$ for every $\mu$.M.L.-random
sequence $x_{1:\infty}$, since {\em every} sequence is $\MM$.M.L.\
random. Since $\MM$ is {\em not} estimable, Vovk's theorem cannot
be applied and it is not obvious how to generalize it. So the
question of individual convergence remains open. More generally,
one may ask whether $\xi_\M\to\mu$ for every $\mu/\xi$-random
sequence. It turns out that this is true for some $\M$, but false for others.

\ftheorem{thMLConv}{$\mu/\xi$-convergence of $\xi$ to $\mu$}{
Let $\X=\B$ be binary and
$\M_\Theta:=\{\mu_\th:\mu_\th(1|x_{<t})=\th\,\forall t,\;
\th\in\Theta\}$ be the set of Bernoulli($\th$) distributions
with parameters $\th\in\Theta$. Let $\Theta_D$ be a countable
dense subset of $[0,1]$, e.g.\ $[0,1]\cap\SetQ$, and let $\Theta_G$
be a countable subset of $[0,1]$ with a gap in the sense that
there exist $0<\th_0<\th_1<1$ such that
$[\th_0,\th_1]\cap\Theta_G=\{\th_0,\th_1\}$, e.g.\
$\Theta_G=\{\odf,\odt\}$ or $\Theta_G=([0,{1\over
4}]\cup[{1\over 2},1])\cap\SetQ$. Then
\begin{list}{}{\ifjournal\itemsep=1ex\fi}
\item[$i)$] If $x_{1:\infty}$ is $\mu/\xi_{\M_{\Theta_D}}$ random with
$\mu\in\M_{\Theta_D}$, then $\xi_{\M_{\Theta_D}}(x_t|x_{<t})\to\mu(x_t|x_{<t})$,
\item[$ii)$] There are $\mu\in\M_{\Theta_G}$ and $\mu/\xi_{\M_{\Theta_G}}\!\!$
random $x_{1:\infty}$ for which
$\xi_{\M_{\Theta_G}}\!\!(x_t|x_{<t})\not\to\mu(x_t|x_{<t})\!\!$
\end{list}
}

\noindent Our original/main motivation of studying
$\mu/\xi$-randomness is the implication of Theorem~\ref{thMLConv}
that $\MM\stackrel{\mbox{\tiny M.L.}}\longrightarrow\mu$ cannot be
decided from $M$ being a mixture distribution or from the
universality property (Theorem~\ref{thUniM}) alone. Further
structural properties of $\M_{enum}^{semi}$ have to be employed.
For Bernoulli sequences, convergence $\mu.\xi_{\M_\Theta}$.r.\ is
related to denseness of $\M_\Theta$. Maybe a denseness
characterization of $\M_{enum}^{semi}$ can solve the question of
convergence M.L.\ of $M$. The property $\MM\in\M_{enum}^{semi}$ is
also not sufficient to resolve this question, since there are
$\M\ni\xi$ for which $\xi\stackrel{\mu.\xi.r}\longrightarrow\mu$
and $\M\ni\xi$ for which
$\xi\not\stackrel{\mu.\xi.r}\longrightarrow\mu$. Theorem
\ref{thMLConv} can be generalized to i.i.d.\ sequences over
general finite alphabet $\X$.

The idea to prove $(ii)$ is to construct a sequence $x_{1:\infty}$
that is $\mu_{\th_0}/\xi$-random {\em and} $\mu_{\th_1}/\xi$-random
for $\th_0\neq\th_1$. This is possible if and only if $\Theta$
contains a gap and $\th_0$ and $\th_1$ are the boundaries of the
gap. Obviously $\xi$ cannot converge to $\th_0$ {\em and} $\th_1$,
thus proving non-convergence. For no $\th\in[0,1]$ will this
$x_{1:\infty}$ be $\mu_\th$ M.L.-random. Finally, the proof of
Theorem~\ref{thMLConv}
makes essential use of the mixture representation of $\xi$, as
opposed to the proof of Theorem~\ref{thConv} which only needs
dominance $\xi\geqm\M$.

An example for $(ii)$ is $\M=\{\mu_0,\mu_1\}$,
$\mu_0(1|x_{<t})=\mu_1(0|x_{<t})={1\over 4}$,
$x_{1:\infty}=(01)^\infty=01010101...$ $\Rightarrow$ $\mu_0(x_{1:2n})=
\mu_1(x_{1:2n})=\xi(x_{1:2n})=({1\over 4})^n({3\over 4})^n$
$\Rightarrow$ $x_{1:\infty}$ is
$\mu_0/\xi$-random {\em and}
$\mu_1/\xi$-random, but
$\mu_0(x_{2n}|x_{<2n})={1\over 4}$,
$\mu_0(x_{2n+1}|x_{1:2n})={3\over 4}$,
$\mu_1(x_{2n}|x_{<2n})={3\over 4}$,
$\mu_1(x_{2n+1}|x_{1:2n})={1\over 4}$ and
$\xi(x_{2n}|x_{<2n})={3\over 8}$,
$\xi(x_{2n+1}|x_{1:2n})={1\over 2}$ for $w_0=w_1=\odt$
$\Rightarrow$ $\xi(x_n|x_{<n})\not\to\mu_{0/1}(x_n|x_{<n})$.

\paradot{Proof}
Let $\X=\B$ and $\M=\{\mu_\th:\th\in\Theta\}$ with countable
$\Theta\subset[0,1]$ and
$\mu_\th(1|x_{1:n})=\th=1-\mu_\th(0|x_{1:n})$, which implies
\beqn
  \mu_\th(x_{1:n}) = \th^{n_1}(1-\th)^{n-n_1},\qquad
  n_1:=x_1\!+...+\!x_n, \qquad
  \hat\th\equiv\hat\th_n:={n_1\over n}
\eeqn
$\hat\th$ depends on $n$; all other used/defined $\th$ will be
independent of $n$. We assume $\th_{\!\cdot\cdot}\in\Theta$, where
$..$ stands for some (possible empty) index, and
$\ddot\th\in[0,1]$ (possibly $\not\in\Theta$), where $\ddot{}$
stands for some superscript, i.e.\ $\mu_{\th_{\!\cdot\cdot}}$ and
$w_{\th_{\!\cdot\cdot}}$ make sense, whereas $\mu_{\ddot\th}$ and
$w_{\ddot\th}$ do not. $\xi$ is defined in the standard way as
\beq\label{MLxiuni}
  \xi(x_{1:n})=\sum_{\th\in\Theta}w_\th\mu_\th(x_{1:n})
  \quad\Rightarrow\quad
  \xi(x_{1:n})\geq w_\th \mu_\th(x_{1:n}),
\eeq
where $\sum_\th w_\th=1$ and $w_\th>0\,\forall\th$.
In the following let $\mu=\mu_{\th_0}\in\M$ be the true environment.
\beq\label{MLmuMr}
  \omega=x_{1:\infty} \mbox{ is } \mu/\xi\mbox{-random}
  \quad\Leftrightarrow\quad
  \exists c_\omega : {\xi(x_{1:n})\leq c_\omega\!\cdot\!\mu_{\th_0}(x_{1:n})}
  \;\forall n
\eeq
For binary alphabet it is sufficient to establish whether
$\xi(1|x_{1:n}) \toinfty{n} \th_0\equiv\mu(1|x_{1:n})$ for
$\mu/\xi$-random $x_{1:\infty}$ in order to decide
$\xi(x_n|x_{<n})\to\mu(x_n|x_{<n})$.
We need the following posterior
representation of $\xi$:
\beq\label{MLpw}
  \xi(1|x_{1:n})=\sum_{\th\in\Theta}w_n^\th \mu_\th(1|x_{1:n}),\quad
  w_n^\th:=w_\th{\mu_\th(x_{1:n})\over\xi(x_{1:n})}
  \leq {w_\th\over w_{\th_0}}{\mu_\th(x_{1:n})\over\mu_{\th_0}(x_{1:n})},\quad
  \sum_{\th\in\Theta}w_n^\th=1
\eeq
The ratio $\mu_\th/\mu_{\th_0}$ can be represented as follows:
\beq\label{MLmuRatio}
  {\mu_\th(x_{1:n})\over\mu_{\th_0}(x_{1:n})}
  = {\th^{n_1}(1\!-\!\th)^{n-n_1}\over \th_0^{n_1}(1\!-\!\th_0)^{n-n_1}}
  = \left[\bigg({\th\over\th_0}\bigg)^{\hat\th_n}
          \bigg({1\!-\!\th\over 1\!-\!\th_0}\bigg)^{1-\hat\th_n}\right]^n
  = \mbox{\Large\e}^{\,\displaystyle n[D(\hat\th_n||\th_0)\!-\!D(\hat\th_n||\th)]}
\eeq
\beqn
  \qmbox{where}\textstyle
  D(\hat\th||\th) = \hat\th\ln{\hat\th\over\th} +
                    (1\!-\!\hat\th)\ln{1-\hat\th\over 1-\th}
\eeqn
is the relative entropy between $\hat\th$ and $\th$, which is
continuous in $\hat\th$ and $\th$, and is $0$ if and only if
$\hat\th=\th$. We also need the following implication for sets
$\Omega\subseteq\Theta$:
\bqa \nonumber
  & & \mbox{If}\quad
  w_n^\th\leq w_\th g_\th(n)\toinfty{n} 0 \qmbox{and}
  g_\th(n)\leq c\;\forall\th\!\in\!\Omega,
\\ \label{MLsumconv}
  & & \mbox{then}\quad
  \sum_{\th\in\Omega}w_n^\th \mu_\th(1|x_{1:n}) \;\leq\;
  \sum_{\th\in\Omega}w_n^\th \toinfty{n} 0,
\eqa
which easily follows from boundedness $\sum_\th w_n^\th\leq 1$ and
$\mu_\th\leq 1$ \cite[Lem.5.28$ii$]{Hutter:04uaibook}. We now
prove Theorem~\ref{thMLConv}. We leave the special considerations
necessary when $0,1\in\Theta$ to the reader and assume,
henceforth, $0,1\not\in\Theta$.

{\bf (i)} Let $\Theta$ be a countable dense subset of $(0,1)$ and
$x_{1:\infty}$ be $\mu/\xi$-random. Using (\ref{MLxiuni}) and
(\ref{MLmuMr}) in (\ref{MLmuRatio}) for $\th\in\Theta$ to be
determined later we can bound
\beq\label{MLenbnd2}
  \e^{n[D(\hat\th_n||\th_0)-D(\hat\th_n||\th)]}
  = {\mu_\th(x_{1:n})\over\mu_{\th_0}(x_{1:n})}
  \leq {c_\omega\over w_\th}
  = :c<\infty
\eeq
Let us assume that $\hat\th\equiv\hat\th_n\not\to\th_0$. This
implies that there exists a cluster point $\tilde\th\neq\th_0$ of
sequence $\hat\th_n$, i.e.\ $\hat\th_n$ is infinitely often in an
$\eps$-neighborhood of $\tilde\th$, e.g.\ $D(\hat\th_n||\tilde\th)\leq\eps$
for infinitely many $n$. $\tilde\th\in[0,1]$ may be outside $\Theta$.
Since $\tilde\th\neq\th_0$ this implies that $\hat\th_n$ must be ``far''
away from $\th_0$ infinitely often. For instance, for $\eps={1\over
4}(\tilde\th-\th_0)^2$, using $D(\hat\th||\tilde\th)+D(\hat\th||\th_0)
\geq (\tilde\th-\th_0)^2$, we get $D(\hat\th||\th_0)\geq 3\eps$. We
now choose $\th\in\Theta$  so near to $\tilde\th$ such that
$|D(\hat\th||\th)-D(\hat\th||\tilde\th)|\leq\eps$ (here we use
denseness of $\Theta$). Chaining all inequalities we get
$D(\hat\th||\th_0)-D(\hat\th||\th)\geq 3\eps-\eps-\eps=\eps>0$.
This, together with (\ref{MLenbnd2}) implies $\e^{n\eps}\leq c$ for
infinitely many $n$ which is impossible. Hence, the assumption
$\hat\th_n\not\to\th_0$ was wrong.

Now, $\hat\th_n\to\th_0$ implies that for arbitrary
$\th\neq\th_0$, $\th\in\Theta$ and for sufficiently large $n$
there exists $\delta_\th>0$ such that $D(\hat\th_n||\th)\geq 2\delta_\th$
(since $D(\th_0||\th)\neq 0)$ and $D(\hat\th_n||\th_0)\leq\delta_\th$.
This implies
\beqn\label{MLwto0}
  w_n^\th \;\leq\; {w_\th\over w_{\th_0}}
  \e^{n[D(\hat\th_n||\th_0)\!-\!D(\hat\th_n||\th)]}
  \;\leq\; {w_\th\over w_{\th_0}} \e^{-n\delta_\th}
  \;\toinfty{n}\; 0,
\eeqn
where we have used (\ref{MLpw}) and (\ref{MLmuRatio}) in the first
inequality and the second inequality holds for sufficiently large
$n$. Hence $\sum_{\th\neq\th_0} w_n^\th\to 0$ by (\ref{MLsumconv})
and $w_n^{\th_0}\to 1$ by normalization (\ref{MLpw}), which finally gives
\beqn
  \xi(1|x_{1:n})=w_n^{\th_0} \mu_{\th_0}(1|x_{1:n}) +
  \sum_{\th\neq\th_0}w_n^\th \mu_\th(1|x_{1:n}) \;\toinfty{n}
  \mu_{\th_0}(1|x_{1:n}).
\eeqn

{\bf (ii)} We first consider the case $\Theta=\{\th_0,\th_1\}$:
Let us choose $\bar\th$ ($=\ln({1-\th_0\over
1-\th_1})/\ln({\th_1\over\th_0}{1-\th_0\over 1-\th_1})
\not\in\Theta$) in the (KL) middle of $\th_0$ and $\th_1$ such
that
\beq\label{MLMid}
  D(\bar\th||\th_0)=D(\bar\th||\th_1), \qquad
  0 < \th_0 < \bar\th < \th_1 < 1,
\eeq
\beqn
  \mbox{and choose $x_{1:\infty}$ such that $\hat\th_n:={n_1\over n}$
  satisfies $|\hat\th_n-\bar\th|\leq{1\over n}
  \quad(\Rightarrow\;\hat\th_n\toinfty{n}\bar\th)$}
\eeqn
We will show that $x_{1:\infty}$
is $\mu_{\th_0}/\xi$-random {\em and} $\mu_{\th_1}/\xi$-random.
Obviously no $\xi$ can converge to $\th_0$
{\em and} $\th_1$, thus proving $\M$-non-convergence.
($x_{1:\infty}$ is obviously not $\mu_{\th_{0/1}}$ M.L.-random,
since the relative frequency $\hat\th_n\not\to\th_{0/1}$.
$x_{1:\infty}$ is not even $\mu_{\bar\th}$ M.L.-random, since
$\hat\th_n$ converges too fast ($\sim\odn$). $x_{1:\infty}$ is
indeed very regular, whereas ${n_1\over n}$ of a truly
$\mu_{\bar\th}$ M.L.-random sequence has fluctuations of the order
$1/\sqrt n$. The fast convergence is necessary for
doubly $\mu/\xi$-randomness.
The reason that $x_{1:\infty}$ is $\mu/\xi$-random, but not M.L.-random is
that $\mu/\xi$-randomness is a weaker concept than M.L.-randomness for
$\M\subset\M_{enum}^{semi}$. Only regularities characterized by
$\nu\in\M$ are recognized by $\mu/\xi$-randomness.)

In the following we assume that $n$ is sufficiently large
such that $\th_0\leq\hat\th_n\leq\th_1$.  We need
\beq\label{MLDD}
  |D(\hat\th||\th)-D(\bar\th||\th)| \leq c|\hat\th-\bar\th|
  \quad\forall\,\th,\hat\th,\bar\th\in[\th_0,\th_1]
  \qmbox{with} \textstyle c:=\ln\!{\th_1(1-\th_0)\over\th_0(1-\th_1)} < \infty
\eeq
which follows for $\hat\th\geq\bar\th$ (similarly
$\hat\th\leq\bar\th$) from
\beqn
  D(\hat\th||\th)-D(\bar\th||\th) = \int_{\bar\th}^{\hat\th}
  [{\textstyle\ln{\th'\over\th}-\ln{1-\th'\over 1-\th}}]d\th'
  \leq \int_{\bar\th}^{\hat\th}
  [{\textstyle\ln{\th_1\over\th_0}-\ln{1-\th_1\over 1-\th_0}}]d\th'
  = c\!\cdot\!(\hat\th-\bar\th)
\eeqn
where we have increased $\th'$ to $\th_1$ and decreased $\th$ to
$\th_0$ in the inequality. Using (\ref{MLDD}) in (\ref{MLmuRatio})
twice we get
\beq\label{MLmu01}
  {\mu_{\th_1}(x_{1:n})\over\mu_{\th_0}(x_{1:n})}
  =
  \e^{n[D(\hat\th_n||\th_0)-D(\hat\th_n||\th_1)]}
  \leq
  \e^{n[D(\bar\th||\th_0)+c|\hat\th_n-\bar\th|-
       D(\bar\th||\th_1)+c|\hat\th_n-\bar\th|]}
  \leq
  \e^{2c}
\eeq
where we have used (\ref{MLMid}) in the last inequality. Now,
(\ref{MLmu01}) and (\ref{MLpw}) lead to
\beq\label{MLwgeq0}
  w_n^{\th_0}
  = w_{\th_0}{\mu_{\th_0}(x_{1:n})\over\xi(x_{1:n})}
  = [1+{w_{\th_1}\over w_{\th_0}}{\mu_{\th_1}(x_{1:n})\over\mu_{\th_0}(x_{1:n})}]^{-1}
  \geq [1+{w_{\th_1}\over w_{\th_0}}\e^{2c}]^{-1}=:c_0>0,
\eeq
which shows that $x_{1:\infty}$ is $\mu_{\th_0}/\xi$-random by
(\ref{MLmuMr}). Exchanging $\th_0\leftrightarrow\th_1$ in
(\ref{MLmu01}) and (\ref{MLwgeq0}) we similarly get
$w_n^{\th_1}\geq c_1>0$, which implies (using
$w_n^{\th_0}+w_n^{\th_1}=1$)
\beq\label{MLnonconv2}
  \xi(1|x_{1:n})=
  \sum_{\th\in\{\th_0,\th_1\}}w_n^\th \mu_\th(1|x_{1:n})
  = w_n^{\th_0}\!\cdot\!\th_0 + w_n^{\th_1}\!\cdot\!\th_1
  \neq \th_0 = \mu_{\th_0}(1|x_{1:n}).
\eeq
This shows $\xi(1|x_{1:n}) \;\;\not\!\!\!\toinfty{n}
\mu(1|x_{1:n})$.
One can show that $\xi(1|x_{1:n})$ does not only not converge to
$\th_0$ (and $\th_1$), but that it does not converge at all. The
fast convergence demand $|\hat\th_n-\bar\th|\leq\odn$ on
$x_{1:\infty}$ can be weakened to
$\hat\th_n\leq\bar\th+O(\odn)\,\forall n$ and
$\hat\th_n\geq\bar\th-O(\odn)$ for infinitely many $n$, then
$x_{1:\infty}$ is still $\mu_{\th_0}/\xi$-random, and
$w_n^{\th_1}\geq c_1'>0$ for infinitely many $n$, which is
sufficient to prove $\xi\not\to\mu$.

We now consider general $\Theta$ with gap in the sense that there exist
$0<\th_0<\th_1<1$ with
$[\th_0,\th_1]\cap\Theta=\{\th_0,\th_1\}$: We show
that all $\th\neq\th_0,\th_1$ give asymptotically no contribution
to $\xi(1|x_{1:n})$, i.e.\ (\ref{MLnonconv2}) still applies. Let
$\th\in\Theta\setminus\{\th_0,\th_1\}$; all other definitions as
before. Then
$\delta_\th:=D(\bar\th||\th)-D(\bar\th||\th_{0/1})>0$, since
$\th$ is farther than $\th_{0/1}$ away from $\bar\th$
($|\th-\bar\th|>|\th_{0/1}-\bar\th|$). Similarly to (\ref{MLmu01}) with
$\th$ instead $\th_1$ we get
\beqn
  {\mu_\th(x_{1:n})\over\mu_{\th_0}(x_{1:n})}
  = \e^{n[D(\hat\th_n||\th_0)-D(\hat\th_n||\th)]}
  \leq \e^{2c}\!\cdot\!
    \e^{n[D(\bar\th||\th_0)-D(\bar\th||\th)]}
  = \e^{2c}\e^{-n\delta_\th}
  \toinfty{n} 0
\eeqn
Hence $w_n^\th\leq{w_\th\over w_{\th_0}}\e^{2c}\e^{-n\delta_\th}\to
0$ from (\ref{MLpw}) and
$\eps_n:=\sum_{\th\in\Theta\setminus\{\th_0,\th_1\}}
w_n^\th\mu_\th(1|x_{1:n})\toinfty{n} 0$ from (\ref{MLsumconv}).
Hence $
  \xi(1|x_{1:n})
  = w_n^{\th_0}\cdot\th_0 + w_n^{\th_1}\cdot\th_1 + \eps_n
  \neq \th_0 = \mu_{\th_0}(1|x_{1:n})
$
for sufficiently large $n$, since $\eps_n\to 0$, $w_n^{\th_1}\geq c'_1>0$
and $\th_0\neq\th_1$.
\qed

\section{Conclusions}\label{secConc}

For a hierarchy of four computability definitions, we completed
the classification of the existence of computable (semi)measures
dominating all computable (semi)measures. Dominance is an important
property of a prior, since it implies rapid convergence of the
corresponding posterior with probability one.
A strengthening would be convergence for all Martin-L{\"o}f (M.L.)
random sequences. This seems natural, since M.L.\ randomness can
be defined in terms of Solomonoff's prior $M$, so there is a close
connection.
Contrary to what was believed before, the question of posterior
convergence $M/\mu\to 1$ for all M.L.\ random sequences is still
open. Some exciting progress has been made recently in
\cite{Hutter:04mlconvx}, partially answering this question.
We introduced a new flexible notion of
$\mu/\xi$-randomness which contains Martin-L{\"of} randomness as a
special case. Though this notion may have a wider range of
application, the main purpose for its introduction was to show
that standard proof attempts of
$M/\mu\stackrel{M.L.}\longrightarrow 1$ based on dominance only
must fail. This follows from the derived result that the validity
of $\xi/\mu\to 1$ for $\mu/\xi$-random sequences depends on the
Bayes mixture $\xi$.


\begin{small}

\end{small}

\begin{thebibliography}{Hut03b}

\bibitem[Cha75]{Chaitin:75}
G.~J. Chaitin.
\newblock A theory of program size formally identical to information theory.
\newblock {\em Journal of the ACM}, 22(3):329--340, 1975.

\bibitem[Doo53]{Doob:53}
J.~L. Doob.
\newblock {\em Stochastic Processes}.
\newblock Wiley, New York, 1953.

\bibitem[G{\'a}c74]{Gacs:74}
P.~G{\'a}cs.
\newblock On the symmetry of algorithmic information.
\newblock {\em Soviet Mathematics Doklady}, 15:1477--1480, 1974.

\bibitem[HM04]{Hutter:04mlconvx}
M.~Hutter and An.~A. Muchnik.
\newblock Universal convergence of semimeasures on individual random sequences.
\newblock In {\em Proc. 15th International Conf. on Algorithmic Learning Theory
  ({ALT-2004})}, volume 3244 of {\em LNAI}, pages 234--248, Padova, 2004.
  Springer, Berlin.

\bibitem[Hut01]{Hutter:01alpha}
M.~Hutter.
\newblock Convergence and error bounds for universal prediction of nonbinary
  sequences.
\newblock In {\em Proc. 12th European Conf. on Machine Learning (ECML-2001)},
  volume 2167 of {\em LNAI}, pages 239--250, Freiburg, 2001. Springer, Berlin.

\bibitem[Hut03a]{Hutter:03unipriors}
M.~Hutter.
\newblock On the existence and convergence of computable universal priors.
\newblock In {\em Proc. 14th International Conf. on Algorithmic Learning Theory
  ({ALT-2003})}, volume 2842 of {\em LNAI}, pages 298--312, Sapporo, 2003.
  Springer, Berlin.

\bibitem[Hut03b]{Hutter:03unimdl}
M.~Hutter.
\newblock Sequence prediction based on monotone complexity.
\newblock In {\em Proc. 16th Annual Conf. on Learning Theory ({COLT-2003})},
  volume 2777 of {\em LNAI}, pages 506--521, Washington, DC, 2003. Springer,
  Berlin.

\bibitem[Hut04]{Hutter:04uaibook}
M.~Hutter.
\newblock {\em Universal Artificial Intelligence: Sequential Decisions based on
  Algorithmic Probability}.
\newblock Springer, Berlin, 2004.
\newblock 300 pages, http://www.idsia.ch/$_{^{\sim}}$marcus/ai/uaibook.htm.

\bibitem[Kol65]{Kolmogorov:65}
A.~N. Kolmogorov.
\newblock Three approaches to the quantitative definition of information.
\newblock {\em Problems of Information and Transmission}, 1(1):1--7, 1965.

\bibitem[Lam87]{Lambalgen:87}
{M. van} Lambalgen.
\newblock {\em Random Sequences}.
\newblock PhD thesis, University of Amsterdam, 1987.

\bibitem[Lev73]{Levin:73random}
L.~A. Levin.
\newblock On the notion of a random sequence.
\newblock {\em Soviet Mathematics Doklady}, 14(5):1413--1416, 1973.

\bibitem[Lev74]{Levin:74}
L.~A. Levin.
\newblock Laws of information conservation (non-growth) and aspects of the
  foundation of probability theory.
\newblock {\em Problems of Information Transmission}, 10(3):206--210, 1974.

\bibitem[LV97]{Li:97}
M.~Li and P.~M.~B. Vit\'anyi.
\newblock {\em An Introduction to {K}olmogorov Complexity and its
  Applications}.
\newblock Springer, Berlin, 2nd edition, 1997.

\bibitem[Sch71]{Schnorr:71}
C.~P. Schnorr.
\newblock {\em Zuf{\"a}lligkeit und Wahrscheinlichkeit}.
\newblock Springer, Berlin, 1971.

\bibitem[Sch00]{Schmidhuber:00toe}
J.~Schmidhuber.
\newblock Algorithmic theories of everything.
\newblock Report IDSIA-20-00, quant-ph/0011122, {IDSIA}, Manno (Lugano),
  Switzerland, 2000.

\bibitem[Sch02]{Schmidhuber:02gtm}
J.~Schmidhuber.
\newblock Hierarchies of generalized {Kolmogorov} complexities and
  nonenumerable universal measures computable in the limit.
\newblock {\em International Journal of Foundations of Computer Science},
  13(4):587--612, 2002.

\bibitem[Sim77]{Simpson:77}
S.~G. Simpson.
\newblock Degrees of unsolvability: A survey of results.
\newblock In J.~Barwise, editor, {\em Handbook of Mathematical Logic}, pages
  631--652. North-Holland, Amsterdam, 1977.

\bibitem[Sol64]{Solomonoff:64}
R.~J. Solomonoff.
\newblock A formal theory of inductive inference: Parts 1 and 2.
\newblock {\em Information and Control}, 7:1--22 and 224--254, 1964.

\bibitem[Sol78]{Solomonoff:78}
R.~J. Solomonoff.
\newblock Complexity-based induction systems: Comparisons and convergence
  theorems.
\newblock {\em IEEE Transaction on Information Theory}, IT-24:422--432, 1978.

\bibitem[VL00]{Vitanyi:00}
P.~M.~B. Vit\'anyi and M.~Li.
\newblock Minimum description length induction, {B}ayesianism, and {K}olmogorov
  complexity.
\newblock {\em IEEE Transactions on Information Theory}, 46(2):446--464, 2000.

\bibitem[Vov87]{Vovk:87}
V.~G. Vovk.
\newblock On a randomness criterion.
\newblock {\em Soviet Mathematics Doklady}, 35(3):656--660, 1987.

\bibitem[Wan96]{Wang:96}
Y.~Wang.
\newblock {\em Randomness and Complexity}.
\newblock PhD thesis, Universit{\"a}t Heidelberg, 1996.

\bibitem[ZL70]{Zvonkin:70}
A.~K. Zvonkin and L.~A. Levin.
\newblock The complexity of finite objects and the development of the concepts
  of information and randomness by means of the theory of algorithms.
\newblock {\em Russian Mathematical Surveys}, 25(6):83--124, 1970.

\end{thebibliography}
\end{document}